\documentclass[pmlr,cleveref]{jmlr} 

\newcommand{\tagdsmode}{nonproceedings}

\makeatletter
\newcommand{\tagdssubmission}{submission}
\newcommand{\tagdsproceedings}{proceedings}

\ifx\tagdsmode\tagdsproceedings

\else\ifx\tagdsmode\tagdssubmission
  \def\ps@jmlrtps{%
    \let\@mkboth\@gobbletwo
    \def\@oddhead{\scriptsize Under Review at the 2nd Conference on Topology, Algebra, and Geometry in Data Science\hfill}%
    \let\@evenhead\@oddhead
    \def\@oddfoot{}%
    \let\@evenfoot\@oddfoot
  }

\else
  \def\ps@jmlrtps{%
    \let\@mkboth\@gobbletwo
    \def\@oddhead{}%
    \let\@evenhead\@oddhead
    \def\@oddfoot{}%
    \let\@evenfoot\@oddfoot
  }
\fi\fi
\makeatother

\usepackage{longtable}

\usepackage{booktabs}
\usepackage[load-configurations=version-1]{siunitx} 

\theorembodyfont{\upshape}
\theoremheaderfont{\scshape}
\theorempostheader{:}
\theoremsep{\newline}

\jmlrvolume{334}
\jmlryear{2026}
\jmlrworkshop{Topology, Algebra, and Geometry in Data Science}

\title[Finsler Geometry, Graph Neural Networks, and You]{Finsler Geometry, Graph Neural Networks, and You}

\ifx\tagdsmode\tagdssubmission

\else

  \author{\Name{T. Mitchell Roddenberry} \Email{mitch@rice.edu} \\
   \Name{Richard G. Baraniuk} \Email{richb@rice.edu}\\
   \addr Rice University, Houston, TX, USA}

\fi

\ifx\tagdsmode\tagdsproceedings
\editor{Editor's name}
\fi

\AddToHook{cmd/appendix/before}{%
    \crefalias{section}{appendix}%
    \crefalias{subsection}{appendix}
}

\usepackage{amsfonts}
\usepackage{tikz-cd}
\usepackage{quiver}
\usepackage[T1]{fontenc}

\usepackage{tikz,pgfplots}
\usepgfplotslibrary{polar}

\usepackage[scaled]{berasans}
\tikzset{every picture/.style={/utils/exec={\sffamily}}}

\usepackage{wrapfig}
\usepackage{caption}
\DeclareMathOperator{\Tr}{Tr}
\renewcommand{\div}{\ensuremath\mathrm{div}}
\newcommand{\id}{\ensuremath\mathrm{id}}

\DeclareMathOperator*{\argmin}{arg\,min}
\DeclareMathOperator*{\diag}{diag}
\DeclareMathOperator*{\rank}{rank}

\usepackage{cite}

\usepackage{autonum}

\usepackage{marginnote}
\usepackage{xcolor}
\newcommand{\todo}[1]{}
\newcommand{\citeX}[1]{}
\newcommand{\richb}[1]{}
\renewcommand{\todo}[1]{\footnote{\color{red} [TODO: {#1}]}}
\renewcommand{\citeX}[1]{{\color{red} [X{#1}]}}
\renewcommand{\richb}[1]{\footnote{\color{green!50!black} [richb: #1]}}

\renewcommand{\hat}[1]{\widehat{#1}}

\begin{document}

\maketitle

\begin{abstract}
  Graph neural network architectures based on the graph Laplacian approximate the Laplace--Beltrami operator, thus limiting their application to isotropic operators.
  As a nonlinear alternative to the Laplace--Beltrami operator, we consider estimates of the Finsler Laplacian on point clouds sampled from a manifold.
  We prove that these discrete estimates converge to the true operator on the manifold as the number of point samples grows.
  Moreover, we show that this operator can be expressed as a graph neural network layer, which we use to define a family of Finslerian graph neural networks constrained to express Finsler geometry.
  We show that Finslerian graph neural networks recover the geometry underlying nonlinear diffusion equations in practice.
\end{abstract}

\section{Introduction}

Graph neural networks operate on discrete collections of points with specified pairwise relations.
These methods incorporate the relations into the network parameterization, often inspired by the ``right'' operator on a limiting object that the graph approximates.
For example, the graph Laplacian is understood as a discrete approximation of the Laplace--Beltrami operator, so that graph neural networks using the Laplacian implicitly treat the graph as being sampled from a Riemannian manifold.

\begin{wrapfigure}{r}{0.37\textwidth}
  \centering
  \includegraphics[width=\linewidth]{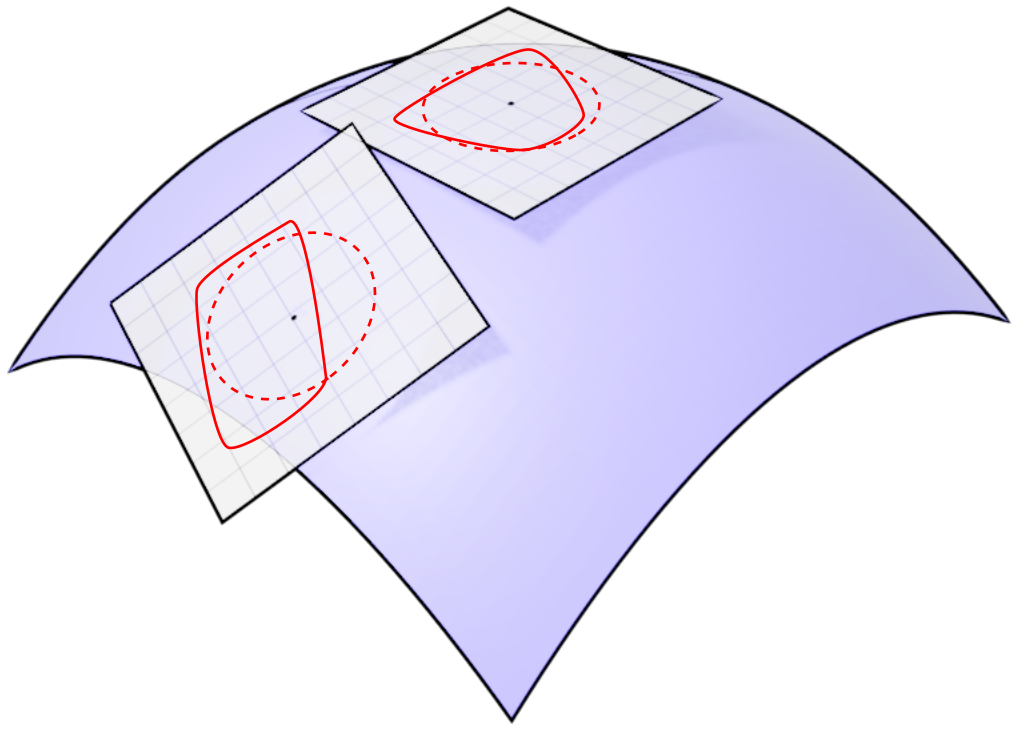}
  \captionsetup{format=plain}
  \caption{Riemannian (dashed) and Finsler (solid) unit balls on the tangent bundle of a manifold. The Finsler metric exhibits anisotropy, non-ellipticity, and asymmetry.}
  \label{fig:minkowski}
\end{wrapfigure}
Using a basic operator, such as the Laplacian, graph neural networks are constructed by interleaving the operator and local nonlinearities, mimicking the composition of affine transforms and nonlinearities in feedforward neural networks.
Despite the nonlinearity, these neural networks are essentially isotropic.

The natural generalization of Riemannian geometry to allow for anisotropy is \emph{Finsler geometry}, which equips the tangent spaces of a smooth manifold with a general Minkowski norm, rather than an inner product.
This allows for asymmetric, non-elliptical geometry, as pictured in \Cref{fig:minkowski}.
Finsler manifolds share many properties with their Riemannian counterparts, including a Laplace operator as the Fr{\'e}chet derivative of an energy functional.
Defining the Laplacian on a Finsler manifold in this way yields a \emph{nonlinear operator}, making it a viable candidate for nonlinear learning techniques that use interpretable geometric structures from the outset.

Extending the approximation of the Laplace--Beltrami operator on submanifolds by the Laplacian of a sampled graph, we consider the problem of discretely approximating nonlinear Laplacians that inherit their structure from Finsler geometries.
To wit, we
\begin{enumerate}
\item Define the empirical Finsler Laplacian for point clouds (\Cref{sec:background:empirical})
\item Prove that the empirical Finsler Laplacian almost surely converges to the continu{\"u}m operator (\Cref{thm:graph-uniform-convergence})

\item Express the empirical Finsler Laplacian as a graph neural network layer (\Cref{sec:gnn}), and define a class of Finslerian graph neural networks (\Cref{sec:gnn:finslerian})
\item Demonstrate the application of the Finslerian graph neural network to the inverse problem of recovering a Finsler metric from observed heat diffusion (\Cref{sec:experiments}).
\end{enumerate}

\noindent{\bf Related Work.}~
The convergence of graph Laplacians to the Laplace--Beltrami operator on manifolds has been studied from multiple angles~\citep{belkin2004,belkin2008,coifman2006,garcia2020} and application settings, particularly in semi-supervised learning~\citep{trillos2018,calder2023}.
The convergence of $p$-Laplacians has also been studied~\citep{slepcev2019}, yielding a class of methods for semi-supervised learning that are nonlinear, but still isotropic.
Additionally, convergence of graph neural networks to manifold neural networks has been considered~\citep{wang2025} using convolution operators based on the Laplace--Beltrami operator.

There have been recent works on applications of Finsler geometry in computer vision and data science, with emphasis on the use of Randers metrics in particular~\citep{weber2024,dages2025,gahtan2026}.
Our discretization of the Finsler Laplacian draws inspiration from methods for solving partial differential equations on point clouds~\citep{liang2013}.
Similar to Randers metrics, ``magnetic Laplacians'' have been used in graph signal processing~\citep{furutani2019} and subsequently for the construction of graph neural networks~\citep{zhang2021,he2022}, with the goal of modeling asymmetric systems on graphs.

\section{Finsler Laplacians}

Let $F$ be a Minkowski norm on $\mathbb{R}^D$, and let a probability measure $\mu$ supported on a closed, compact, smooth, $d$-dimensional submanifold $\mathcal{M}\subset\mathbb{R}^D$ be given.
Let $p$ be the density of $\mu$ with respect to the Hausdorff measure of the manifold.
Under this structure, $\mathcal{M}$ is a weighted Finsler manifold \citep{bao2000introduction} with metric pulled back from $F$ via the inclusion map, with the co-Finsler metric given by the dual norm $F^*$.
For technical background, we refer to \Cref{sec:finsler}.

Define an energy functional for differentiable functions on $\mathcal{M}$:
\begin{equation}
  \label{eq:cont-energy}
  E[f] = \int \frac{1}{2} \left[ F^*(\nabla f(x)) \right]^2 d\mu(x).
\end{equation}
The \emph{Finsler Laplacian} \citep{ge2000,ohta2009} is a nonlinear operator defined as the Fr{\'e}chet derivative of $E$, or equivalently as the divergence of the gradient of the squared (co-)norm:
\begin{equation}
  \label{eq:cont-fins-laplacian-energy}
  \Delta[f] = \frac{1}{p}D\{E[f]\} = \frac{1}{p}\div\left(p J(\nabla f)\right),
\end{equation}
where
\begin{equation}
  \label{eq:J}
  J(\xi) = \nabla_{\xi}\frac{1}{2}\left[ F^*(\xi) \right]^2
  = F^{*}(\xi)\nabla_{\xi}F^{*}(\xi).
\end{equation}
Observe that taking $F$ as the Euclidean norm recovers the usual density-weighted Laplace--Beltrami operator.
\begin{remark}
    For the reader who likes keeping track of indices, $J:T^*\mathcal{M}\to T\mathcal{M}$ is the Legendre transformation for the cotangent/tangent bundles of the manifold.
\end{remark}

\begin{figure}
  \centering
  \includegraphics[height=0.28\linewidth]{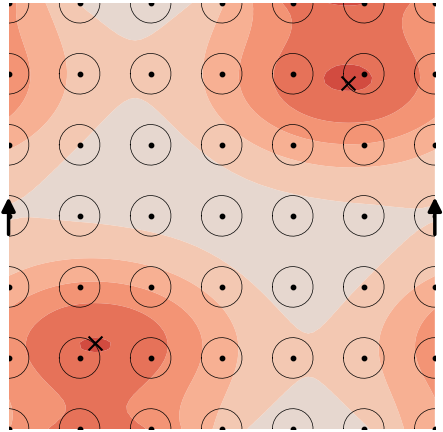}
  \includegraphics[height=0.28\linewidth]{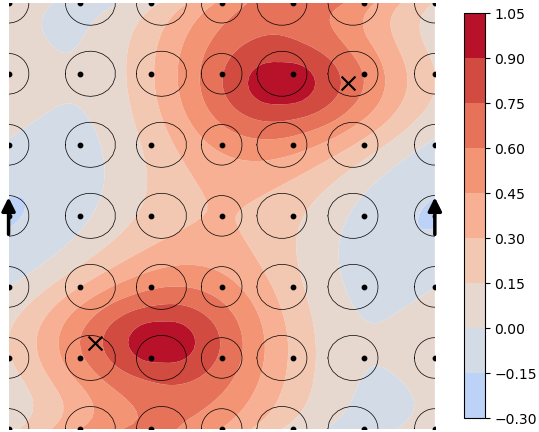}
  \caption{Heat diffusion on a cylinder (identified edges marked by arrows) with two point sources (marked $\times$) using the isotropic Laplace--Beltrami operator (left) and the Finsler Laplacian derived from an ambient Randers metric (right). Panels show Tissot's indicatrices for the respective metrics. The Randers heat equation exhibits ``drift'' in accordance with the off-center geometry.}
  \label{fig:heat}
\end{figure}

\begin{example}
  \label{exam:randers}
  Let $\mathcal{M}\subset\mathbb{R}^3$ be a cylindrical surface.
  We consider the heat equation $\dot{f}(t)=-\Delta[f(t)]$, where $f(0)$ is the sum of two Dirac delta functions on $\mathcal{M}$, and $\Delta$ is either the Laplace--Beltrami operator, or the Finsler Laplacian induced by a Minkowski norm on the ambient space.
  In particular, for some vector $v$ with norm less than one, let $F^{*}(\xi)=\|\xi\|+\langle \xi, v\rangle$, so that $F^{*}$ is a Randers metric on the cotangent bundle of the ambient space.
  As shown in \Cref{fig:heat}, the Randers metric causes the diffusion of heat to ``drift'' according to the vector $v$.
\end{example}

\subsection{The Finsler Graph Laplacian}
\label{sec:background:empirical}

Let $\mathcal{X}=\{x_i\}_{i=1}^n$ be sampled 
\textit{i.i.d.}\ from the distribution $\mu$, and suppose we have access to the function values $f(x_i)$ for each $i$.
We wish to estimate the gradients $\nabla f$ based on these samples, given a translation-invariant nonnegative kernel $\kappa$ and $\epsilon>0$.
Assume that $\kappa:\mathbb{R}^D\to\mathbb{R}$ is smooth, compactly supported, and radially symmetric.
Define $\kappa_{\epsilon}(z):=\epsilon^{-d}\kappa(z/\epsilon)$.

We use a local principal component analysis method for estimating the gradient of $f$.
Namely, we define
\begin{equation}
  \label{eq:disc-grad-est-loose}
  \nabla_{\epsilon,n} f(x) =
  C_{\epsilon,n}^{\dagger}(x) b_{\epsilon,n}[f](x),
\end{equation}
where $C_{\epsilon,n}^{\dagger}$ is (almost) a pseudoinverse of the kernel-weighted covariance centered at $x$, and $b_{\epsilon,n}[f](x)$ is the kernel-weighted local variation of $f$ near $x$.\footnote{We offer more precise definitions of these objects in \Cref{sec:gradient}.}
Using the empirical gradient, we define the empirical energy functional
\begin{equation}
  \label{eq:disc-energy}
  E_{\epsilon,n}[f] = \frac{1}{n}\sum_{i=1}^n
  \frac{1}{2}\left[ F^*(\nabla_{\epsilon,n} f(x_i)) \right]^2,
\end{equation}
and the \emph{empirical Finsler Laplacian} via the Fr{\'e}chet derivative: $\Delta_{\epsilon,n}[f] = n D \{ E[f] \}$.
For convenience, define the (co)vector $\xi(x)=C_{\epsilon,n}^{\dagger}(x)J(\nabla_{\epsilon,n}f(x))$.
This yields the formula
\begin{equation}
  \label{eq:disc-fins-laplacian-ptwise-2}
  \Delta_{\epsilon,n}[f](x)
  = \frac{1}{n\epsilon^{2}}\sum_{i=1}^n
  \kappa_{\epsilon}\left(x-x_i\right)
  \langle \xi(x) + \xi(x_i), x-x_i \rangle.
\end{equation}
We define the \emph{Finsler graph Laplacian} $\mathcal{L}$ as the restriction of the empirical Finsler Laplacian to the point cloud $\mathcal{X}$.


Although the Finsler Laplacian and its graph counterpart are derived from similar energy functionals, the capacity of the Finsler graph Laplacian to approximate the underlying operator does not follow immediately.
Since $n\to\infty$, if $\epsilon\to 0$ at a suitable rate, the Finsler graph Laplacian converges to the continu{\"u}m Finsler Laplacian in the following sense:
\begin{theorem}
  \label{thm:graph-uniform-convergence}
  Let $\mu$ be a probability measure supported on a smooth, closed, and compact $d$-dimensional submanifold $\mathcal{M}\subset\mathbb{R}^D$, with density $p$ bounded away from zero on its support.
  Assume that $f$ is a $C^3$ function on this submanifold.
  For each $n\geq 1$, let $\{x_i\}_{i=1}^n$ be a set of points sampled \textit{i.i.d.}\ according to the measure $\mu$.
  Then, for $\epsilon = O \left( \log n / n \right)^{1/(3d+4)}$,
  \begin{equation}
    \label{eq:graph-uniform-bound}
    \lim_{n\to\infty}\max_{j\in[1,n]}\Big|
    \Delta[f](x_j) - \mathcal{L}[f](x_j)
    \Big| = 0 \quad\text{almost surely}.
  \end{equation}
\end{theorem}
We prove this in \Cref{sec:proof-ptwise}.

\section{The Finsler Graph Laplacian and Neural Networks}
\label{sec:gnn}

Cellular sheaves are a powerful tool for organizing operators on graphs that transform data between vector spaces~\citep{curry2014,hansen2020sheaf,barbero2022sheaf}.
We cast the Finsler graph Laplacian in this framework, which motivates a new graph neural network architecture.

\subsection{Cellular Sheaves for the Finsler Laplacian}

Let $\mathcal{X}=\{x_i\}_{i=1}^n$ be a point cloud, with an observed function $f:\mathcal{X}\to\mathbb{R}$.
After constructing a graph using $\mathcal{X}$, we construct sheaf structures over the graph such that appropriate composition of restriction maps, morphisms, and their adjoints computes the Finsler graph Laplacian.
Construct a graph $\mathcal{G}=(\mathcal{X},\mathcal{E})$ so that $\mathcal{X}$ is the set of nodes and $\mathcal{E}$ consists of unordered pairs of nodes $(x_i,x_j)$ such that $\kappa_\epsilon(x_i-x_j)\neq 0$.
We define two cellular sheaves\footnote{See \Cref{sec:app-sheaf} for background information on cellular (co)sheaves.} on $\mathcal{G}$.
The first we denote by $\mathcal{F}$, such that all vector spaces $\mathcal{F}(x)$ and $\mathcal{F}(e)$ are the real line $\mathbb{R}$.
For any incident node-edge pair, the restriction map is defined as $\mathcal{F}_{x\to e}f_x = f_x$
for $f_x\in \mathcal{F}(x)\simeq\mathbb{R}$.
The second sheaf is denoted by $\mathcal{T}$, and assigns a copy of the tangent space $T_x\mathbb{R}^D$ to each node $x\in \mathcal{X}$, while still assigning the real line to each edge $e\in \mathcal{E}$.
For each node $x\in \mathcal{X}$ and any incident edge $(x,y)\in \mathcal{E}$, the restriction map is
\begin{equation}
  \label{eq:tangent-restriction}
  \mathcal{T}_{x\to (x,y)}v = \frac{1}{n\epsilon^{2}}
  \kappa_{\epsilon} \left(x-y\right)
  \langle C_{\epsilon,n}^{\dagger}(x) (x-y), v\rangle,
\end{equation}
for $v\in T_x\mathbb{R}^D\simeq\mathbb{R}^D$.
We identify the tangent spaces with the cotangent spaces $T_x\mathbb{R}^D\simeq T_x^*\mathbb{R}^D$ using the ambient Euclidean metric.
We define the following maps on the spaces of edge/node data:
\begin{itemize}
\item $\id: C^1(\mathcal{F})\to C^{1}(\mathcal{T})$ and $\id^*: C^1(\mathcal{T})\to C^1(\mathcal{F})$ are both identity maps under the obvious identification $C^1(\mathcal{F})\simeq C^1(\mathcal{T})$,
\item $J: C^0(\mathcal{T})\to C^0(\mathcal{T})$ passes (co)vector data on the nodes through the map $J$ as defined in \eqref{eq:J}.
\end{itemize}
The vector spaces constituting both sheaves are endowed with the usual inner product structure; 
the restriction maps and their adjoints induce a differential $d:C^0\to C^1$ and a codifferential $\delta:C^1\to C^0$ for both sheaves $\mathcal{F},\mathcal{T}$.
We model the function on the nodes as a $0$-cochain $f\in C^0(\mathcal{F})$.
The Finsler graph Laplacian is represented as
\begin{equation}
  \label{eq:sheaf-finsler}
  \mathcal{L}[f] = (\delta \circ \id^{*} \circ d \circ J \circ \delta \circ \id \circ d)[f],
\end{equation}
where $d:C^0\to C^1$ generically indicates the differential and $\delta:C^1\to C^0$ the codifferential.
Perhaps more legibly, the Finsler graph Laplacian traverses the following (non-commutative) diagram:
\begin{equation}
  \label{eq:sheaf-finsler-diagram}
  \begin{tikzcd}
    {C^0(\mathcal{F})} & {C^1(\mathcal{F})} & {C^1(\mathcal{T})} & {C^0(\mathcal{T})}
    \arrow["d", shift left, from=1-1, to=1-2]
    \arrow["\delta", shift left, from=1-2, to=1-1]
    \arrow["\id", shift left, from=1-2, to=1-3]
    \arrow["{\id^*}", shift left, from=1-3, to=1-2]
    \arrow["\delta", shift left, from=1-3, to=1-4]
    \arrow["d", shift left, from=1-4, to=1-3]
    \arrow["J", shift left, from=1-4, to=1-4, loop, in=325, out=35, distance=10mm]
  \end{tikzcd}.
\end{equation}
%

Considering the map $J$ as an activation function, the Finsler graph Laplacian can be thought of as a layer in a sheaf neural network~\citep{hansen2020sheaf,barbero2022sheaf}, where the nonlinearity is applied in a ``lifted'' space (namely, the cotangent space at each point).
In contrast, the nonlinearity for a typically defined sheaf neural network is applied in the same domain as the input space, that is, as a map $\sigma:C^0(\mathcal{F})\to C^0(\mathcal{F})$.

\subsection{Finslerian Graph Neural Networks}
\label{sec:gnn:finslerian}

We now consider how a graph neural network may be designed to parameterize a Finsler metric.
Specifically, we will replace $J$ with a simple neural network.
For $W,V\in\mathbb{R}^{E\times D}$, and an activation function $\sigma:\mathbb{R}\to\mathbb{R}$ that is positive homogeneous of degree $1$ with $\sigma'\geq 0$, define the approximation
\begin{equation}
  \label{eq:approximate-J}
  \hat{J}(\xi) =
  V^\top\sigma(W\xi),
\end{equation}
where $\sigma$ is applied elementwise in the standard basis. 
The following result gives a condition that guarantees $\hat{J}$ to be ``Finslerian,'' in the sense of being the gradient of a squared-norm:
\begin{proposition}
  \label{prop:symmetry}
  If $V=W$ in \eqref{eq:approximate-J}, $\rank(W)=D$, and $\sigma'>0$, then $\hat{J}(\xi)=\nabla_{\xi}\frac{1}{2} \left[ \hat{F}^{*}(\xi) \right]^2$ where
  \begin{equation}
    \label{eq:approximate-E}
    \hat{F}^{*}(\xi)
    =
    \sqrt{2\langle \xi, \hat{J}(\xi) \rangle}
  \end{equation}
  is an (almost) Minkowski norm.
  Under the further assumption that $\sigma(x)=\alpha x$ for $\alpha>0$, the norm $\hat{F}^{*}$ is Riemannian.
\end{proposition}
We prove \Cref{prop:symmetry} in \Cref{sec:finsler:symmetry}.
Based on this, for a (potentially constant) learnable matrix $W(t)$, we define a \emph{Finslerian graph neural network} under the framework of graph neural diffusion~\citep{chamberlain2021,thorpe2022,bodnar2022neural};
that is, an ordinary differential equation of the form
\begin{equation}
  \label{eq:finsler-gnn}
  \dot{f}(t) = -\hat{\mathcal{L}}_{W(t)}[f(t)],
\end{equation}
where $\hat{\mathcal{L}}_{W(t)}$ is the Finsler graph Laplacian defined using $\hat{J}(\xi)=W^{\top}(t)\sigma(W(t)\xi)$.

\section{Numerical Experiments}
\label{sec:experiments}

We demonstrate the Finsler graph Laplacian and Finslerian graph neural networks on a collection of numerical examples, corroborating our convergence result (\Cref{thm:graph-uniform-convergence}) and the utility of a symmetric nonlinear network for representing Minkowski norms (\Cref{prop:symmetry}).
Implementation details can be found in \Cref{sec:methods}, and source code is available 
\href{https://github.com/tmrod/finsler-gnn}{[here]}.

\subsection{Convergence of the Finsler Graph Laplacian}

\begin{figure}
  \centering
  \includegraphics[height=0.2\linewidth]{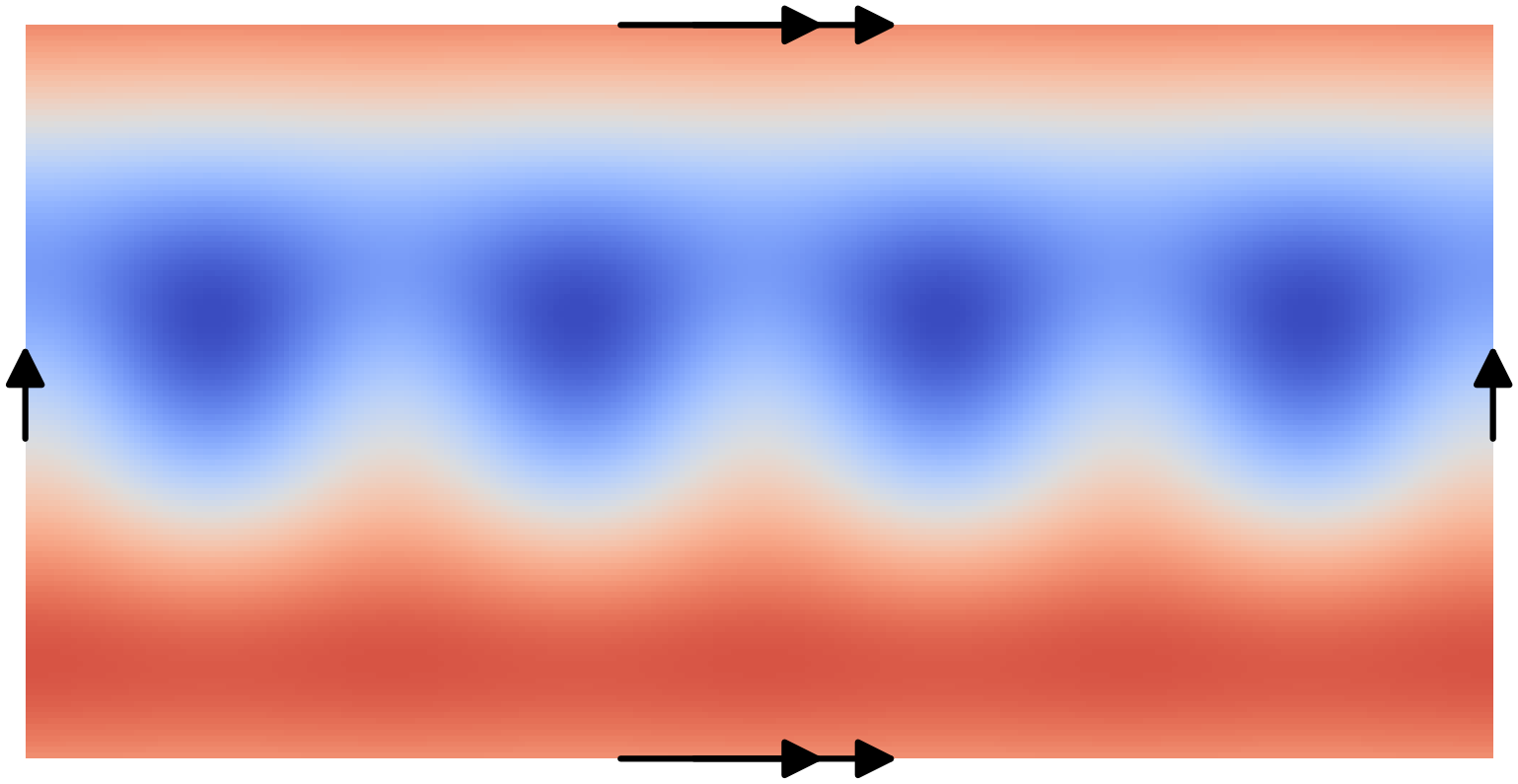}
  \includegraphics[height=0.2\linewidth]{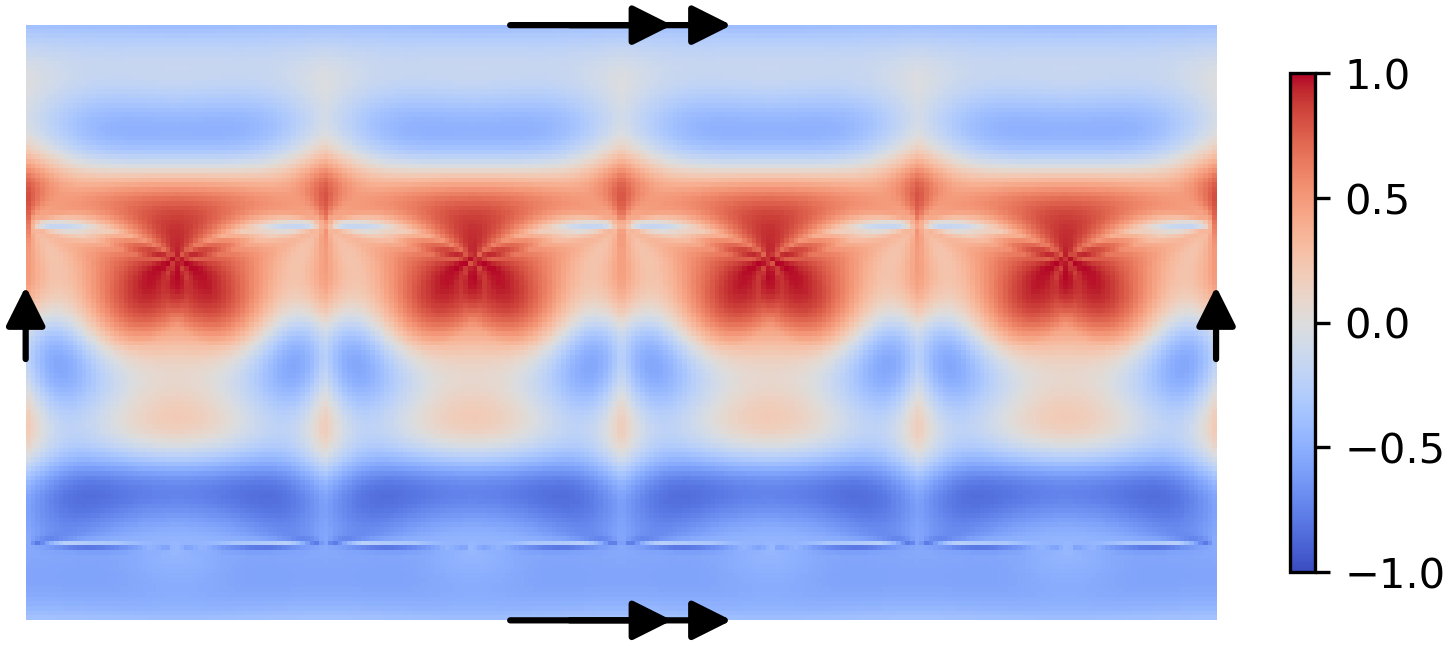}
  \caption{The function $f(x,y,z)=\cos(x)+\cos(y)+z$ restricted to the torus (left) and the Finsler Laplacian $\Delta[f]$ defined so that $F^{*}(\xi)=\|\xi\|_3$ (right). Identified edges of torus marked by arrows.}
  \label{fig:convergence-functions}
\end{figure}

We first validate the convergence result stated in \Cref{thm:graph-uniform-convergence}.
Let $\mathcal{M}\subset\mathbb{R}^3$ be a torus with major radius $2$ and minor radius $1$ oriented along the $xy$-plane.
We consider the restriction of a function $f(x,y,z)=\cos(x)+\cos(y)+z$ on the ambient space to $\mathcal{M}$, and consider the Finsler Laplacian defined by the $\ell_3$-norm, that is, where $F^{*}(\xi)= \left( \sum_i |\xi_i|^3 \right)^{1/3}$.
The function $f$ and the application of the Finsler Laplacian $\Delta[f]$ are pictured in \Cref{fig:convergence-functions}.
\begin{figure}
  \centering
  \includegraphics[width=0.5\linewidth]{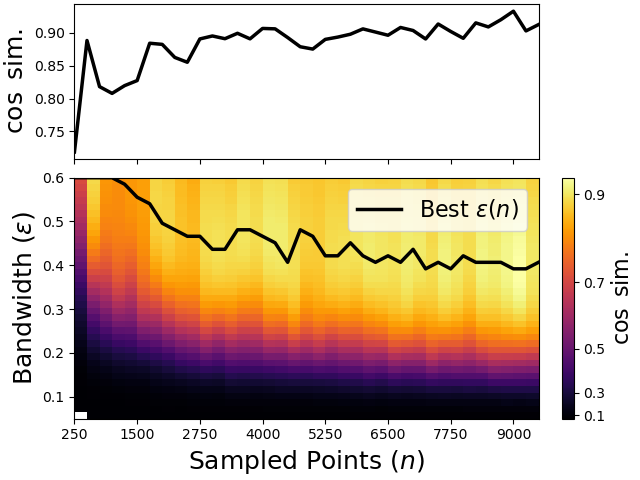}
  \begin{tikzpicture}
  \begin{polaraxis}[
    x axis line style = {draw=none},
    width=0.41\linewidth,
    xticklabel=\empty,
    yticklabel style={anchor=north, yshift=0.195\linewidth},
    y axis line style={yshift=0.2\linewidth},
    ytick style={yshift=0.19\linewidth},
    legend style={
      at={(-0.15,1.15)},
      anchor=north west,
      font=\footnotesize
    }
    ]
    \addplot [data cs=polarrad, no markers, thick, dashed, black]
    table [col sep=comma, x=theta, y expr=sqrt(\thisrow{rtrue})]
    {data/exp8_002_ball.csv};
    \addlegendentry{G.T.};
    \addplot [data cs=polarrad, no markers, thick, black]
    table [col sep=comma, x=theta, y expr=sqrt(\thisrow{rmodel})]
    {data/exp8_002_ball.csv};
    \addlegendentry{ReLU};
    \addplot [data cs=polarrad, no markers, thick, blue]
    table [col sep=comma, x=theta, y expr=sqrt(\thisrow{rmodel})]
    {data/exp8_004_ball.csv};
    \addlegendentry{Linear};
  \end{polaraxis}
\end{tikzpicture}

  \caption{(Bottom left) Similarity between the Finsler graph Laplacian and the true Finsler Laplacian as a function of $n$ and $\epsilon$. (Top left) Cosine similarity for ``best'' $\epsilon(n)$. (Right) Recovery of a Randers metric by a Finslerian graph neural network.}
  \label{fig:experiments}
\end{figure}
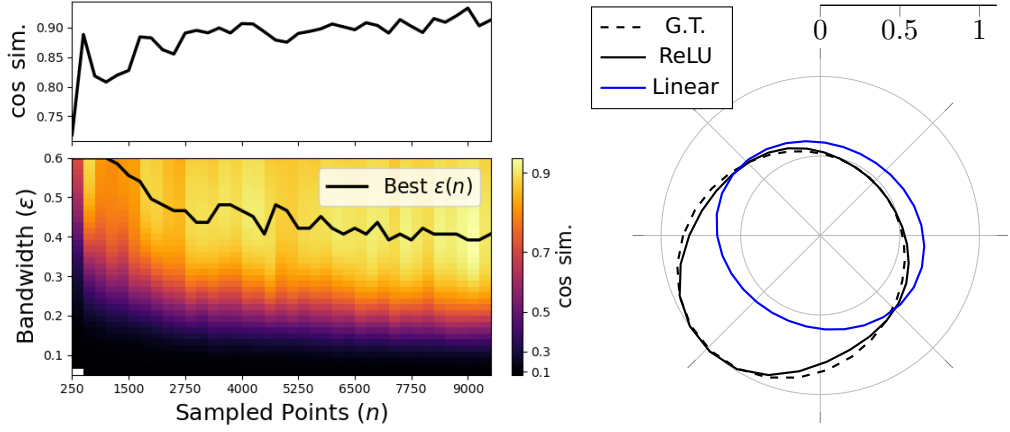
We sample points on the torus uniformly according to the ambient (Hausdorff) measure, and evaluate the Finsler graph Laplacian constructed using varying numbers of sampled points $n$ and bandwidths $\epsilon$.
We then measure the cosine similarity between the true Finsler Laplacian evaluated at the sampled points and the Finsler graph Laplacian, to account for scaling.
The results of this are shown in \Cref{fig:experiments}~(left).
As expected, the similarity between the true and graph Laplacian map increases as $n$ grows from $250$ to $9500$, with the corresponding ``best'' value of $\epsilon$ decreasing at a modest rate as the point cloud becomes denser.
This corroborates the convergence result established by \Cref{thm:graph-uniform-convergence}.

\subsection{Learning Geometry using Finslerian Graph Neural Networks}

We consider a semisupervised geometric inverse problem of recovering a Finsler metric from an observed instance of heat diffusion.
Namely, we sample a point cloud of $n=600$ points uniformly on the unit square, and construct a graph using a gaussian kernel with a bandwidth of $\epsilon=0.2$.
In the graph construction, we periodize the domain, giving the manifold the topology of a torus.
We generate a function $f(0)$ on the domain as a superposition of $5$ Fourier modes of varying frequency, phase, and amplitude, then sample it to yield a function on the point cloud.
Then, we run the heat equation using a Finsler Laplacian defined using a Randers (co)metric (see \Cref{exam:randers}) with a drift vector $v=[0.25,0.25]$.
We observe the state $f(T)$ for $T=1$, and use $20\%$ of the labeled nodes for training.

Given this single input-output sample of the heat equation over one time unit, we apply multiple techniques to model the heat diffusion.
The first two are Finslerian graph neural networks (FGNNs), both with hidden dimension $E=64$, where one uses the ReLU activation and the other uses a linear activation.
The third and fourth learn maps on the node signal space that more closely resembles a graph neural network (GNN). 
The initial input $f(0)$ is linearly lifted to $\mathbb{R}^E$, and heat diffusion is performed for node features in $\mathbb{R}^E$ where $\dot{f}(t)=\mathrm{MLP}(Lf(t))$ with $L$ denoting the standard graph Laplacian, and $\mathrm{MLP}$ a single-hidden-layer neural network with either ReLU or linear activations, mapping $\mathbb{R}^E$ to $\mathbb{R}^E$ on each node.
The output of the diffusion is then linearly projected back to $\mathbb{R}$.

We train all methods to map the initial input $f(0)$ to predict the labeled $20\%$ of nodes at $f(T)$.
We then measure the relative mean-squared error (MSE) over the remaining $80\%$ of nodes, recorded under the column ``Train Graph'' in \Cref{tab:mse}.
The Finslerian graph neural networks strongly outperform the standard graph neural networks.
\begin{wraptable}{r}{0.52\textwidth}
  \centering
  \captionsetup{format=plain}
  \caption{%
    MSE of learned approximations to Finsler heat flow.
  }
  \label{tab:mse}
  \begin{tabular}{lrr}
    \toprule
    Method & Train Graph & Test Graph \\
    \midrule
    \textbf{ReLU FGNN}    & $\mathbf{1.6\times 10^{-5}}$ & $\mathbf{1.1\times 10^{-5}}$ \\
    Linear FGNN  & $4.6\times 10^{-4}$          & $1.4\times 10^{-4}$ \\
    \midrule
    ReLU GNN & $1.9\times 10^{-2}$          & $1.5\times 10^{-1}$ \\
    Linear GNN & $2.2\times 10^{-2}$          & $1.8\times 10^{-1}$ \\
    \bottomrule
\end{tabular}
\end{wraptable}
Moreover, the nonlinear (ReLU) Finslerian graph neural network performs an order of magnitude better than the linear one, because the linear method is only capable of \emph{Riemannian} diffusion, by \Cref{prop:symmetry}.
To test generalization, we resample the graph and generate a new initial condition $f(0)$, and compare the trained models to the true Finsler heat equation on the new, unseen graph.
As shown in \Cref{tab:mse} under the column ``Test Graph,'' the Finslerian graph neural networks suffer no performance loss, reflecting the generalization suggested by \Cref{thm:graph-uniform-convergence}.
On the other hand, the graph neural networks appear to overfit to the training graph, performing much worse on the test graph.

We also examine the qualitative properties of the Finslerian graph neural networks after training.
Since both methods amounts to learning $\hat{J}(\xi)=\nabla_{\xi}\frac{1}{2}[ \hat{F}^{*}(\xi)]^2$, we plot the unit balls of the norms $\hat{F}^{*}$ in \Cref{fig:experiments}~(right).
The only approach that yields a metric close to the ground truth (G.T.) metric is the model with ReLU activation.
The model with linear activation learns a metric whose unit ball is an ellipse, again due to its restriction to Riemannian metrics by \Cref{prop:symmetry}.

\section{Conclusion}

We have developed the {topological} and {algebraic} tools for approximating Finsler {geometric} Laplacian operators on sampled point clouds.
The proposed empirical Finsler Laplacian is shown to converge in a uniform pointwise sense, demonstrating its utility both for computations in PDEs and in semisupervised learning problems arising in {data science}.
Beyond the applications, the empirical Finsler Laplacian demonstrates the limitations of graph signal processing and graph neural network methods that rely too heavily on isotropy.
By linking the structure of the Finsler Laplacian to a class of nonlinear graph neural networks, we hope to promote study into how nonlinearities in neural architectures implicitly impose geometric structures on the domain.
By tying the neural architecture design to a known geometric structure, such as a Minkowski norm, solutions found by learning algorithms become interpretable via the induced geometry.

\acks{This work was supported by ONR grant N00014-23-1-2714, DOE grant DE-SC0020345, DOI grant 140D0423C0076, and a Google Cloud Computing Award.}

\bibliography{references}

@article{burago1993,
  title={Isometric embeddings of {Finsler} manifolds},
  author={Burago, Dmitri and Ivanov, Sergey},
  journal={Algebra i Analiz},
  volume=5,
  number=1,
  year=1993,
  note={English: St. Petersburg Math Journal 5.1 (1994)}
}

@book{bao2000introduction,
  title={An {Introduction} to {Riemann-Finsler} Geometry},
  author={Bao, David and Chern, S-S and Shen, Zhongmin},
  year=2000,
  publisher={Springer Science \& Business Media}
}

@article{ge2000,
  title={Eigenvalues and eigenfunctions of metric measure manifolds},
  author={Ge, Yuxin and Shen, Zhongmin},
  journal={Proceedings of the London Mathematical Society},
  volume=82,
  number=3,
  year=2000
}

@article{ohta2009,
  title={Heat flow on {Finsler} manifolds},
  author={Ohta, Shin-ichi and Sturm, Karl-Theodor},
  journal={Communications on Pure and Applied Mathematics},
  volume=62,
  number=10,
  year=2009
}

@article{liang2013,
  title={Solving partial differential equations on point clouds},
  author={Liang, Jian and Zhao, Hongkai},
  journal={SIAM Journal on Scientific Computing},
  volume=35,
  number=3,
  year=2013
}

@article{davis2026,
  title={Characteristic tensors for almost {Finsler} manifolds},
  author={Davis, James F and Edwards, Benjamin R and Kosteleck{\`y}, V Alan},
  journal={The Journal of Geometric Analysis},
  volume=36,
  number=4,
  year=2026
}

@article{trillos2018,
  title={A variational approach to the consistency of spectral clustering},
  author={Trillos, Nicol{\'{a}}s Garc{\'{i}}a and Slep{\v{c}}ev, Dejan},
  journal={Applied and Computational Harmonic Analysis},
  volume=45,
  number=2,
  year=2018,
}

@article{slepcev2019,
  title={Analysis of {$p$-Laplacian} regularization in semisupervised learning},
  author={Slep{\v{c}}ev, Dejan and Thorpe, Matthew},
  journal={SIAM Journal on Mathematical Analysis},
  volume=51,
  number=3,
  year=2019
}

@article{garcia2020,
  title={Error estimates for spectral convergence of the graph {Laplacian} on random geometric graphs toward the {Laplace--Beltrami} operator},
  author={Trillos, Nicol{\'a}s Garc{\'\i}a and Gerlach, Moritz and Hein, Matthias and Slep{\v{c}}ev, Dejan},
  journal={Foundations of Computational Mathematics},
  volume=20,
  number=4,
  year=2020
}

@article{calder2023,
  title={Rates of convergence for {Laplacian} semi-supervised learning with low labeling rates},
  author={Calder, Jeff and Slep{\v{c}}ev, Dejan and Thorpe, Matthew},
  journal={Research in the Mathematical Sciences},
  volume=10,
  number=1,
  pages=10,
  year=2023
}

@InProceedings{wang2025,
  title = 	 {A Manifold Perspective on the Statistical Generalization of Graph Neural Networks},
  author =       {Wang, Zhiyang and Cervino, Juan and Ribeiro, Alejandro},
  booktitle = 	 {International Conference on Machine Learning},
  year = 	 2025
}

@inproceedings{furutani2019,
  title={Graph signal processing for directed graphs based on the {Hermitian Laplacian}},
  author={Furutani, Satoshi and Shibahara, Toshiki and Akiyama, Mitsuaki and Hato, Kunio and Aida, Masaki},
  booktitle={Joint European Conference on Machine Learning and Knowledge Discovery in Databases},
  year=2019
}

@inproceedings{zhang2021,
  title={{MagNet}: A neural network for directed graphs},
  author={Zhang, Xitong and He, Yixuan and Brugnone, Nathan and Perlmutter, Michael and Hirn, Matthew},
  booktitle={Advances in Neural Information Processing Systems},
  year=2021
}

@inproceedings{he2022,
  title={{MSGNN}: A spectral graph neural network based on a novel magnetic signed {Laplacian}},
  author={He, Yixuan and Perlmutter, Michael and Reinert, Gesine and Cucuringu, Mihai},
  booktitle={Learning on Graphs Conference},
  year=2022,
}

@inproceedings{weber2024,
  title={{Finsler-Laplace-Beltrami} operators with application to shape analysis},
  author={Weber, Simon and Dages, Thomas and Gao, Maolin and Cremers, Daniel},
  booktitle={IEEE/CVF Conference on Computer Vision and Pattern Recognition},
  year=2024
}

@inproceedings{dages2025,
  title={Finsler multi-dimensional scaling: Manifold learning for asymmetric dimensionality reduction and embedding},
  author={Dages, Thomas and Weber, Simon and Lin, Ya-Wei Eileen and Talmon, Ronen and Cremers, Daniel and Lindenbaum, Michael and Bruckstein, Alfred M and Kimmel, Ron},
  booktitle={IEEE/CVF Conference on Computer Vision and Pattern Recognition},
  year=2025
}

@article{gahtan2026,
  title={Wildfire Simulation with Differentiable {Randers-Finsler} Eikonal Solvers},
  author={Gahtan, Barak and Shpund, Jacob and Bronstein, Alex M},
  journal={arXiv:2603.00035},
  year=2026
}

@inproceedings{chamberlain2021,
  title={{GRAND}: Graph neural diffusion},
  author={Chamberlain, Ben and Rowbottom, James and Gorinova, Maria I and Bronstein, Michael and Webb, Stefan and Rossi, Emanuele},
  booktitle={International Conference on Machine Learning},
  year=2021
}

@inproceedings{thorpe2022,
  title={{GRAND++}: Graph neural diffusion with a source term},
  author={Thorpe, Matthew and Nguyen, Tan Minh and Xia, Hedi and Strohmer, Thomas and Bertozzi, Andrea and Osher, Stanley and Wang, Bao},
  booktitle={International Conference on Learning Representations},
  year=2022
}

@inproceedings{hansen2020sheaf,
  title={Sheaf Neural Networks},
  author={Jakob Hansen and Thomas Gebhart},
  booktitle={NeurIPS Workshop on Topological Data Analysis and Beyond},
  year=2020
}

@article{bodnar2022neural,
  title={Neural sheaf diffusion: A topological perspective on heterophily and oversmoothing in {GNNs}},
  author={Bodnar, Cristian and Di Giovanni, Francesco and Chamberlain, Benjamin and Li{\`o}, Pietro and Bronstein, Michael},
  journal={Advances in Neural Information Processing Systems},
  volume=35,
  year=2022
}

@inproceedings{barbero2022sheaf,
  title={Sheaf neural networks with connection {Laplacians}},
  author={Barbero, Federico and Bodnar, Cristian and de Oc{\'a}riz Borde, Haitz S{\'a}ez and Bronstein, Michael and Veli{\v{c}}kovi{\'c}, Petar and Li{\`o}, Pietro},
  booktitle={Topological, Algebraic and Geometric Learning Workshops},
  year=2022
}

@book{curry2014,
  title={Sheaves, cosheaves and applications},
  author={Curry, Justin Michael},
  year=2014,
  publisher={University of Pennsylvania}
}

@article{belkin2004,
  title={Semi-supervised learning on {Riemannian} manifolds},
  author={Belkin, Mikhail and Niyogi, Partha},
  journal={Machine learning},
  volume=56,
  number=1,
  year=2004
}

@article{belkin2008,
  title={Towards a theoretical foundation for {Laplacian-based} manifold methods},
  author={Belkin, Mikhail and Niyogi, Partha},
  journal={Journal of Computer and System Sciences},
  volume=74,
  number=8,
  year=2008
}

@article{coifman2006,
  title={Diffusion maps},
  author={Coifman, Ronald R and Lafon, St{\'e}phane},
  journal={Applied and Computational Harmonic Analysis},
  volume=21,
  number=1,
  year=2006
}

@software{jax2018,
  author = {James Bradbury and others},
  title = {{JAX}: composable transformations of {P}ython+{N}um{P}y programs},
  url = {http://github.com/jax-ml/jax},
  year = 2018,
}

@article{kidger2021a,
    author={Patrick Kidger and Cristian Garcia},
    title={{E}quinox: neural networks in {JAX} via callable {P}y{T}rees and filtered transformations},
    year=2021,
    journal={Differentiable Programming workshop at Neural Information Processing Systems}
}

@software{deepmind2020,
  title = {The {D}eep{M}ind {JAX} {E}cosystem},
  author = {DeepMind and others},
  url = {http://github.com/google-deepmind},
  year = 2020,
}

@phdthesis{kidger2021b,
    title={{O}n {N}eural {D}ifferential {E}quations},
    author={Patrick Kidger},
    year=2021,
    school={University of Oxford},
}

@article{tsitouras2011,
  title={{Runge--Kutta} pairs of order 5 (4) satisfying only the first column simplifying assumption},
  author={Tsitouras, Charalampos T.},
  journal={Computers \& Mathematics with Applications},
  volume=62,
  number=2,
  year=2011
}

\appendix

\section{Finsler Geometry}
\label{sec:finsler}

We treat the ambient space $\mathbb{R}^D$ as an affine space, so that the tangent spaces $T_x\mathcal{M}$ are all isomorphic to $\mathbb{R}^D$, and the ambient space has no curvature.
A \emph{Minkowski norm} on $\mathbb{R}^D$ is a continuous function $F:\mathbb{R}^D\to\mathbb{R}^{\geq 0}$ with the following properties for all $v,w\in\mathbb{R}^D,\lambda\geq 0$:
\begin{description}
    \item[Smoothness:] $F$ is smooth away from zero.
    \item[Strong convexity:] $\nabla^2[F^2(v)] \succ 0$ for $v\neq 0$.
    \item[Positive homogeneity:] $F(\lambda v)=\lambda F(v)$.
    \item[Positive definiteness:] $F(v)>0$ unless $v=0$.
\end{description}
Unlike a typical norm, we only require homogeneity to hold for $\lambda\geq 0$, rather than for all $\lambda\in\mathbb{R}$; this allows for $F$ to be asymmetric.

If the conditions for the Minkowski norm $F$ only hold on a subset $\mathbb{R}^D\setminus S$ where $S$ is a closed, conic subset of $\mathbb{R}^D$, we say that $F$ is an \emph{almost Minkowski norm}~\citep[Definition~4]{davis2026}.

For a linear subspace $U\subset\mathbb{R}^D$, we pullback the Minkowski norm to $U$ via the inclusion map $\id:U\hookrightarrow\mathbb{R}^D$ as $F(u):= (F\circ\id)(u)$ for $u\in U$.
One can check that this defines a Minkowski norm on $U$.
For a smooth submanifold $\mathcal{M}\subset\mathbb{R}^D$, endowing the tangent bundle of $\mathcal{M}$ with the pullback of the Minkowski norm on $\mathbb{R}^D$ defines a \emph{Finsler manifold}~\citep{bao2000introduction}.

Notably, as shown by \citet[Theorem~3.2]{burago1993}, if the Minkowski norm $F$ is odd and $\mathcal{M}$ is compact, then there exists a Minkowski norm $F'$ on $\mathbb{R}^{2d}$ such that $\mathcal{M}$ can be isometrically embedded in $\mathbb{R}^{2d}$.

\subsection{Proof of \Cref{prop:symmetry}}
\label{sec:finsler:symmetry}

For $\hat{J}$ to be the gradient of a scalar-valued energy functional on the cotangent bundle, the Jacobian of $\hat{J}$ must be a symmetric tensor, since that computes the Hessian of the functional.
Specifically,
\begin{equation}
    \label{eq:jacobian-J}
    \nabla\hat{J}(\xi)
    =
    W^\top \diag(\sigma'(W\xi)) W.
\end{equation}
Observe that enforcing $V=W^{\top}$ guarantees that $\nabla\hat{J}(\xi)$ is symmetric.
Moreover, if $\rank(W)=D$ and $\sigma'>0$, it also holds that the Jacobian of $\hat{J}$ is positive definite.
Moreover, since $\sigma$ is positive homogeneous of degree $1$, the norm \eqref{eq:approximate-E} is also positive homogeneous of degree $1$.
If $\sigma$ is nonsmooth at zero (for instance, if $\sigma$ is the Leaky ReLU), then we only have smoothness of $F$ at points where none of the entries of $W\xi$ are equal to zero: the set of such $\xi$ is a closed, conic subset of $\mathbb{R}^D$ with measure zero.
Hence, we have established the conditions for $\hat{F}^*$ to be an almost Minkowski norm, as desired.

\section{Empirical Gradient Estimation}
\label{sec:gradient}

We develop a suitable estimate of the gradient of a function $f$ given samples over a finite set of points $\{x_i\}_{i=1}^n$.
A first approach is the moving least-squares estimate:
\begin{equation}
  \label{eq:disc-grad-est}
  \nabla_{\epsilon,n} f(x) =
  \argmin_\xi
  \sum_{i=1}^n \kappa_\epsilon\left(x-x_i\right)
  \left( f(x) - f(x_i) - \langle \xi, x-x_i \rangle \right)^2.
\end{equation}
In other words,
\begin{equation}
  \label{eq:disc-grad-est-2}
    \nabla_{\epsilon,n} f(x)
    =
    \argmin_\xi G_{\epsilon,n}(x,\xi),
\end{equation}
where
\begin{equation}
  \label{eq:disc-grad-obj}
  G_{\epsilon,n}(x,\xi) =
  \frac{1}{\epsilon^{2}}\sum_{i=1}^n
  \kappa_\epsilon\left(x-x_i\right)
  \left( f(x) - f(x_i) - \langle \xi, x-x_i \rangle \right)^2.
\end{equation}
Setting $\nabla_\xi G_{\epsilon,n}(x,\xi)=0$ yields the solution
\begin{equation}
  \label{eq:disc-grad-est-3}
  \begin{gathered}
    \nabla_{\epsilon,n} f(x) =
    \left( C_{\epsilon,n}(x) \right)^{-1} b_{\epsilon,n}[f](x) \\
    C_{\epsilon,n}(x) =
    \frac{1}{n\epsilon^{2}}\sum_{i=1}^n 
    \kappa_\epsilon\left(x-x_i\right)
    (x-x_i)\otimes (x-x_i) \\
    b_{\epsilon,n}[f](x) =
    \frac{1}{n\epsilon^{2}}\sum_{i=1}^n 
    \kappa_\epsilon\left(x-x_i\right)
    (f(x)-f(x_i)) (x-x_i).
  \end{gathered}
\end{equation}
We call $\nabla_{\epsilon,n}$ the empirical gradient, $C_{\epsilon,n}$ the empirical covariance, and $b_{\epsilon,n}$ the empirical cogradient.

To account for this, we define
\begin{equation}
  \label{eq:disc-grad-est-4}
  \nabla_{\epsilon,n} f(x) =
  h_{\epsilon} \left( C_{\epsilon,n}(x) \right) b_{\epsilon,n}[f](x)
\end{equation}
where
\begin{equation}
  \label{eq:filter}
  h_{\epsilon}(A) =
  (A^4+\epsilon^4I)^{-1}A^3.
\end{equation}
This choice of $h_{\epsilon}$ approximates the matrix inverse for parts of the spectrum far from zero, while remaining close to zero for parts of the spectrum close to zero;
in other words, it acts as a soft Moore-Penrose pseudoinverse.
Hence, we denote $C_{\epsilon,n}^\dagger(x) := h_\epsilon(C_{\epsilon,n}(x))$.

\begin{remark}
We show in \Cref{sec:proof-ptwise} that the covariance matrices $C_{\epsilon},C_{\epsilon,n}$ concentrate near weighted projection matrices onto local tangent planes, yielding a rank-$d$ structure.
In practice, one could replace $h_{\epsilon}$ with a spectral thresholding function, or with a rank-$d$ approximation.
\end{remark}

\section{Cellular (Co)sheaf Laplacians}
\label{sec:app-sheaf}

We provide background on cellular (co)sheaves.
For deeper coverage, we suggest the thesis of \citet{curry2014}.
A \emph{cellular sheaf} $\mathcal{F}$ on the graph $\mathcal{G}$ assigns
\begin{itemize}
\item to each node $x\in \mathcal{X}$
  a vector space $\mathcal{F}(x)$,
\item to each edge $e\in \mathcal{E}$
  a vector space $\mathcal{F}(e)$, and
\item to each incident vertex-edge pair $(x,e)$ a linear \emph{restriction map} $\mathcal{F}_{x\to e}: \mathcal{F}(x)\to \mathcal{F}(e)$.
\end{itemize}
The \emph{dual cosheaf} $\mathcal{F}^{*}$ on the graph $\mathcal{G}$ assigns
\begin{itemize}
\item to each node $x\in \mathcal{X}$
  the dual vector space $\mathcal{F}^{*}(x)$,
\item to each edge $e\in \mathcal{E}$
  the dual vector space $\mathcal{F}^{*}(e)$, and
\item to each incident vertex-edge pair $(x,e)$ a linear \emph{extension map} $\mathcal{F}^{*}_{x\to e}: \mathcal{F}^{*}(e)\to \mathcal{F}^{*}(x)$.
\end{itemize}

For a given cellular (co)sheaf $\mathcal{F}$, the space of $0$-cochains $C^0(\mathcal{F})$ constitutes the set of all assignments of vectors to each node according to the sheaf structure.
That is, a $0$-cochain $f\in C^0(\mathcal{F})$ is a map sending each $x\in\mathcal{X}$ to an element of $\mathcal{F}(x)$.

The space of $1$-cochains is defined similarly, with additional information to account for the orientation of edges.
A $1$-cochain $\beta\in C^1(\mathcal{F})$ is a map sending each \emph{oriented edge} $\overrightarrow{e}$ to an element of $\mathcal{F}(e)$, with the property that reversing the orientation changes the sign: $\beta(\overleftarrow{e})=-\beta(\overrightarrow{e})$.

This yields the differential map from $C^0(\mathcal{F})$ to $C^1(\mathcal{F})$ for a sheaf $\mathcal{F}$ as follows.
Let $\overrightarrow{e}=[x,y]$ be an oriented edge.
Then, for $f\in C^0(\mathcal{F})$, put
\begin{equation}
    \label{eq:sheaf-differential}
    (df)(\overrightarrow{e}) = \mathcal{F}_{y\to e}(f(y))-\mathcal{F}_{x\to e}(f(x)).
\end{equation}
Similarly, for the dual cosheaf $\mathcal{F}^*$, we have the codifferential map from $C^1(\mathcal{F}^*)$ to $C^0(\mathcal{F}^*)$ defined for $\beta\in C^1(\mathcal{F}^*)$ as
\begin{equation}
    (\delta\beta)(x) = 
    \sum_{y\in\mathcal{X}:(x,y)\in\mathcal{E}}
    \mathcal{F}^*_{x\to e}(\beta([x,y])).
\end{equation}

We now repeat the construction of the Finsler Laplacian.
Let $\mathcal{F}$ and $\mathcal{T}$ be the respective cellular sheaves of \emph{functions} and \emph{vectors} defined in \cref{sec:gnn}, with corresponding dual cosheaves $\mathcal{F}^*$ and $\mathcal{T}^*$ of \emph{cofunctions} and \emph{covectors}.
Let $\id:C^1(\mathcal{F})\to C^1(\mathcal{T}^*),\id^*:C^1(\mathcal{T})\to C^1(\mathcal{F}^*)$ be the identity maps under the obvious identification of the spaces.
Given that $J$ maps covectors to vectors, it can be extended to a map $J:C^0(\mathcal{T}^*)\to C^0(\mathcal{T})$.
Then, we have that the Finsler graph Laplacian is a map $\mathcal{L}:C^0(\mathcal{F})\to C^0(\mathcal{F}^*)$ defined by composition
\begin{equation}
    \label{eq:fancy-graph-finsler}
    \mathcal{L}[f] = (\delta \circ \id^* \circ d \circ J \circ \delta \circ \id \circ d)[f],
\end{equation}
or, as a diagram,
\begin{equation}
    \label{eq:fancy-graph-finsler-diagram}
    \begin{tikzcd}
	{C^0(\mathcal{F})} & {C^0(\mathcal{T}^*)} & {C^0(\mathcal{T})} & {C^0(\mathcal{F}^*)} \\
	{C^1(\mathcal{F})} & {C^1(\mathcal{T}^*)} & {C^1(\mathcal{T})} & {C^1(\mathcal{F}^*)}
	\arrow["d", from=1-1, to=2-1]
	\arrow["J", from=1-2, to=1-3]
	\arrow["d", from=1-3, to=2-3]
	\arrow["\id", from=2-1, to=2-2]
	\arrow["\delta", from=2-2, to=1-2]
	\arrow["{\id^*}", from=2-3, to=2-4]
	\arrow["\delta", from=2-4, to=1-4]
    \end{tikzcd}.
\end{equation}
Of course, if the vector spaces for the sheaves are inner product spaces, we can identify them with their duals.
This allows for the simplification of \eqref{eq:fancy-graph-finsler-diagram} to the diagram \eqref{eq:sheaf-finsler-diagram}.

\section{Proof of Laplacian Convergence}
\label{sec:proof-ptwise}

In this section, we prove a collection of results towards \Cref{thm:graph-uniform-convergence}.
Throughout, we will assume that $\epsilon$ is sufficiently small so that it is less than the reach of the manifold $\mathcal{M}\subset\mathbb{R}^D$.
For fixed $\epsilon>0$, it will be useful to consider the empirical Finsler Laplacian in the limiting r{\'e}gime $n\to\infty$.
Under mild conditions on the kernel $\kappa$ and density $p$, and assuming $f$ is appropriately bounded, we have
\begin{equation}
  \label{eq:smooth-grad-2}
  \begin{gathered}
    \nabla_{\epsilon} f(x) =
    h_{\epsilon}\left( C_{\epsilon}(x) \right)
    \left( b_{\epsilon}[f](x) \right) \\
    C_{\epsilon}(x) =
    \frac{1}{\epsilon^{d+2}}
    \int 
    \kappa\left(\frac{x-y}{\epsilon}\right)
    (x-y)\otimes (x-y) d\mu(y) \\
    b_{\epsilon}[f](x) =
    \frac{1}{\epsilon^{d+2}}
    \int
    \kappa\left(\frac{x-y}{\epsilon}\right)
    (f(x)-f(y)) (x-y) d\mu(y).
  \end{gathered}
\end{equation}

Using this smoothed estimator of the gradient $\nabla_{\epsilon} f$, we define the smoothed asymptotic energy functional
\begin{equation}
  \label{eq:smooth-energy}
  E_{\epsilon}[f] = \int \frac{1}{2}\left[ F^*(\nabla_{\epsilon} f(x)) \right]^2 d\mu(x),
\end{equation}
and the smoothed Laplacian by taking the Fr{\'e}chet derivative
\begin{equation}
  \label{eq:smooth-fins-laplacian}
  \Delta_{\epsilon}[f] = \frac{1}{p}D E_{\epsilon}[f].
\end{equation}
Under the assumption that the kernel $\kappa$ is radially symmetric (isotropic with respect to the Euclidean norm) about the origin, we evaluate the smoothed Laplacian \eqref{eq:smooth-fins-laplacian} to
\begin{equation}
  \label{eq:smooth-fins-laplacian-2}
  \begin{aligned}
    \Delta_{\epsilon}[f](x)
    &=
      \frac{1}{\epsilon^{d+2}}
      \int\kappa\left(\frac{x-y}{\epsilon}\right)
      \Big(
      h_{\epsilon}(C_{\epsilon}(x)) J(\nabla_{\epsilon} f(x))  \\
    &\qquad\qquad\qquad\qquad+
      h_{\epsilon}(C_{\epsilon}(y)) J(\nabla_{\epsilon} f(y)) 
      \Big)
      \cdot
      (x-y)
      d\mu(y).
  \end{aligned}
\end{equation}

The strategy will amount to a bound between the difference of $\Delta_{\epsilon,n}[f]$ and $\Delta[f]$, intermediated by $\Delta_{\epsilon}[f]$ via the triangle inequality.
In particular, we will bound the sum
\begin{equation}
  \label{eq:rough-bound-strategy}
  |\Delta[f] - \Delta_{\epsilon,n}[f]|
  \leq
  |\Delta[f] - \Delta_{\epsilon}[f]|
  + |\Delta_{\epsilon}[f] - \Delta_{\epsilon,n}[f]|.
\end{equation}
The first term measures the approximation error due to the smoothing parameter $\epsilon$, and the second measures the error due to discretization.

\subsection{Local Kernel Moments}

Let $x \in \mathcal{M}$ be given.
Through a rigid transformation, we choose coordinates such that $x=0$ and the tangent plane at $x$ occupies the first $d$ coordinates, with the normal plane occupying the remaining $D-d$ coordinates.
Denoting the tangent coordinates by $u\in T_x \mathcal{M}$, we represent $\mathcal{M}$ locally as a graph $y_x(u) = (u, g_x(u)) \in \mathbb{R}^D$.
The normal coordinates $g_x(u)$ satisfy
\begin{equation}
  \label{eq:normal-coordinates}
  g_x(u) =
  \Pi_x[u,u] + D_x[u,u,u] + O(u^4),
\end{equation}
where $\Pi_x$ is a symmetric tensor given by the second fundamental form at $x$, $D_x$ is a third-order tensor, and $O(u^4)$ denotes some fourth-order tensor expression of $u$.
Denoting the tensor product of a vector $v$ with itself $k$ times by $v^{\otimes k}$, we consider expressions of the form
\begin{equation}
  \label{eq:kernel-moments}
  M(x,k,\epsilon) = \frac{1}{\epsilon^{d+2}}
  \int
  \kappa \left( \frac{y_x(u)}{\epsilon} \right)
  y_x(u)^{\otimes k}
  du.
\end{equation}
First, we carry out the change of variables $v=u/\epsilon$, yielding
\begin{equation}
  \label{eq:kernel-moments-2}
  M(x,k,\epsilon) = \frac{1}{\epsilon^{2}}
  \int
  \kappa \left( \frac{y_x(\epsilon v)}{\epsilon} \right)
  y_x(\epsilon v)^{\otimes k}
  dv.
\end{equation}
We assume that $\kappa(z)$ is a smooth, compactly supported function of $\|z\|^2$, \textit{i.e.}, $\kappa(z)=\kappa(\|z\|^2)$.
By a Taylor series expansion centered at a given $z$, for $w\perp z$, we have
\begin{equation}
  \label{eq:kernel-taylor}
  \begin{aligned}
    \kappa(z + w) &= \kappa(\|z\|^2 + \|w\|^2) \\
    &= \kappa(\|z\|^2) + \frac{\kappa'(\|z\|^2)}{2}\|w\|^2 + O(\|w\|^4).
  \end{aligned}
\end{equation}

\subsubsection{Flattening of the Graph}

With the preliminary definitions handled, we begin by considering the behavior of $\frac{y(\epsilon v}{\epsilon}$ as $\epsilon\to 0$.
Observe that $\frac{y(\epsilon v)}{\epsilon}=(v, g_x(\epsilon v)/\epsilon)$, so the main quantity of interest is $g_x(\epsilon v)$.
By direct substitution into \eqref{eq:normal-coordinates}, we have
\begin{equation}
  \label{eq:scaled-normal-coordinates}
  g_x(\epsilon v)
  =
  \epsilon^2 \Pi_x[v,v] + \epsilon^3 D_x[v,v,v] + \epsilon^4 O(v^4).
\end{equation}
Note that $g_x(\epsilon v) \perp v$, and the squared-norm satisfies
\begin{equation}
  \label{eq:scaled-normal-norm}
  \|g_x(\epsilon v)\|^2
  =
  \epsilon^4 \|\Pi_x[v,v]\|^2 + 2\epsilon^5 \langle \Pi_x[v,v], D_x[v,v,v]\rangle + \epsilon^6 O(v^6).
\end{equation}
We conclude, by substitution into \eqref{eq:kernel-taylor}, that
\begin{equation}
  \label{eq:scaled-kernel-taylor}
  \kappa \left( \frac{y(\epsilon v)}{\epsilon} \right) =
  \kappa(\|v\|^2) +
  \kappa'(\|v\|^2)
  \left( \epsilon^2\|\Pi_x[v,v]\|^2 +
    2\epsilon^3\langle \Pi_x[v,v], D_x[v,v,v]\rangle\right) +
  O(\epsilon^4),
\end{equation}
recalling that the compact support of $\kappa$ allows us to assume $v^6=O(1)$.
That is, $\kappa(y(\epsilon v)/\epsilon)=\kappa(\|v\|^2)+O(\epsilon^2)$, so that the normal coordinates become negligible as $\epsilon\to 0$.

\subsubsection{Parity of Local Kernel Moment Integrals}

Let $T:(T_x \mathcal{M})^k\to\mathbb{R}$ be a multilinear form for some odd value of $k$, and consider the integral of $\kappa(\|v\|^2)T[v,\ldots,v]$.
Since $k$ is odd, $T[v,\ldots,v]$ is odd, so the integral vanishes by radial symmetry, \textit{i.e.},
\begin{equation}
  \label{eq:kernel-odd-parity}
  \int \kappa(\|v\|^2) T[v,\ldots,v] dv = 0.
\end{equation}
A similar property holds for $\kappa'$, so that
\begin{equation}
  \label{eq:kernel-prime-odd-parity}
  \int \kappa'(\|v\|^2) T[v,\ldots,v] dv = 0.
\end{equation}

We define a few quantities of interest for integrals of tensor forms with even parity.
The first is the second moment of $\kappa(\|v\|^2)$:
\begin{equation}
  \label{eq:kernel-second-moment-constant}
  m_{\kappa}I_d = \int \kappa(\|v\|^2) v^{\otimes 2} dv,
\end{equation}
where $I_d$ denotes the $d\times d$ identity matrix.
The second is defined based on the curvature at $x$:
\begin{equation}
  \label{eq:kernel-local-curvature-moment}
  m_{\kappa,x} = \int \kappa(\|v\|^2) \Pi_x[v,v] dv.
\end{equation}
Note that $m_{\kappa}$ is a scalar, but $m_{\kappa,x}$ is a vector normal to $x$.

\subsubsection{Computation of Local Kernel Moments}

With these tools in place, computing the moments $M(x,k,\epsilon)$ becomes a trivial exercise.
When $k=1$, substituting \eqref{eq:scaled-kernel-taylor} and \eqref{eq:scaled-normal-coordinates} into \eqref{eq:kernel-moments-2} and zeroing all terms with odd parity, we have
\begin{equation}
  \label{eq:kernel-first-moment}
  M(x,1,\epsilon) =
  (0, m_{\kappa,x}) + O(\epsilon^2).
\end{equation}
Similarly, when $k=2$, we have
\begin{equation}
  \label{eq:kernel-second-moment}
  M(x,2,\epsilon) =
  \begin{bmatrix}
    m_{\kappa}I_d & 0 \\
    0 & 0
  \end{bmatrix} +
  O(\epsilon^2).
\end{equation}

\subsection{Smoothed Gradient}

We consider now the deviation between the smoothed gradient $\nabla_{\epsilon} f$ and the true gradient $\nabla f$.
As before, let $x\in \mathcal{M}$ be given, and apply a rigid transformation so that $x=0$ and the tangent plane occupies the first $d$ coordinates.
We pull back the density $p$ and the function $f$ to $T_x \mathcal{M}$ so that the relevant integrals may locally be defined as integrals over the tangent plane.
We rewrite \eqref{eq:smooth-grad-2} as
\begin{equation}
  \label{eq:smooth-grad-2-alt}
  \begin{gathered}
    \nabla_{\epsilon} f(x) =
    h_{\epsilon}\left( C_{\epsilon}(x) \right)
    b_{\epsilon}[f](x) \\
    C_{\epsilon}(x) =
    \frac{1}{\epsilon^{d+2}}
    \int_{T_x \mathcal{M}} 
    \kappa\left(\frac{y_{x}(u)}{\epsilon}\right)
    y_x(u)^{\otimes 2} p_x(u) du \\
    b_{\epsilon}[f](x) =
    \frac{1}{\epsilon^{d+2}}
    \int_{T_x \mathcal{M}}
    \kappa\left(\frac{y_x(u)}{\epsilon}\right)
    y_x(u)
    (f_x(u)-f_x(0)) p_x(u) du.
  \end{gathered}
\end{equation}

\subsubsection{Local Taylor Series}

It will be useful to gather some local approximations of $p_x$ and $f_x$ by their Taylor series representations.
Taking a Taylor expansion of $p_x$, we have
\begin{equation}
  \label{eq:density-taylor}
  p_x(u) = p_x(0)
  + \langle \nabla p_x(0), u\rangle
  + \frac{1}{2} \nabla^2 p_x(0)[u,u]
  + \frac{1}{6} \nabla^3 p_x(0)[u,u,u]
  + O(u^4).
\end{equation}
Similarly, carrying out a Taylor series expansion of $f_x(u)-f_x(0)$ centered at zero, we have
\begin{equation}
  \label{eq:function-taylor}
  f_x(u) - f_x(0) =
  \nabla f_x(0)[u]
  + \frac{1}{2} \nabla^2 f_x(0)[u,u]
  + \frac{1}{6} \nabla^3 f_x(0)[u,u,u]
  + O(u^4).
\end{equation}

\subsubsection{Smoothed Covariance}

Performing the substitution $v=u/\epsilon$ in \eqref{eq:smooth-grad-2-alt} yields
\begin{equation}
  \label{eq:smooth-covariance-1}
  C_{\epsilon}(x) =
  \frac{1}{\epsilon^2}
  \int_{T_x \mathcal{M}} 
  \kappa\left(\frac{y_x(\epsilon v)}{\epsilon}\right)
  y_x(\epsilon v)^{\otimes 2} p_x(\epsilon v) dv.
\end{equation}
Substituting in the Taylor series expansion of the density \eqref{eq:density-taylor} yields
\begin{equation}
  \label{eq:smooth-covariance-2}
  C_{\epsilon}(x) =
  \begin{bmatrix}
    p_x(0)m_\kappa I_d  & 0 \\
    0 & 0
  \end{bmatrix}
  + O(\epsilon^2).
\end{equation}
That is, $C_{\epsilon}(x)$ is approximately $p(x)m_\kappa I_d$ in the tangent coordinates, apart from an $O(\epsilon^{2})$ error term.

\subsubsection{Smooth Cogradient}

We carry out the same substitution as before, $v=u/\epsilon$, yielding
\begin{equation}
  \label{eq:smooth-cograd-1}
    b_{\epsilon}[f](0) =
    \frac{1}{\epsilon^2}
    \int
    \kappa\left(\frac{y(\epsilon v)}{\epsilon}\right)
    y(\epsilon v)
    (f_x(\epsilon v)-f_x(0)) p_x(\epsilon v) dv.
\end{equation}
Substituting the Taylor series expansions \eqref{eq:density-taylor} and \eqref{eq:function-taylor}, we have
\begin{equation}
  \label{eq:smooth-cograd-2}
  b_{\epsilon}[f](x)
  = C_{\epsilon}(x)\nabla f(x) + O(\epsilon^2).
\end{equation}

\subsubsection{Combined Error}

By \eqref{eq:smooth-cograd-2}, and using $h_{\epsilon}$ from \eqref{eq:filter}, we have
\begin{equation}
  \label{eq:smooth-grad-3}
  \begin{aligned}
    \nabla_{\epsilon} f(x)
    &= h_{\epsilon}(C_{\epsilon}(x)) C_{\epsilon}(x)\nabla f(x)
    + O(\epsilon^2) h_{\epsilon}(C_{\epsilon}(x)) \\
    &= \nabla f(x) + O(\epsilon),
  \end{aligned}
\end{equation}
where $\nabla f(x)\in\mathbb{R}^D$ by the obvious embedding of the tangent plane.
It follows that $J(\nabla_{\epsilon}f(x)) = J(\nabla f(x)) + O(\epsilon)$.

Since $p(x)$ is bounded away from zero, for sufficiently small $\epsilon>0$ we have $h_{\epsilon}(p(x)m_{\kappa}I_d)\approx \frac{1}{p(x)m_{\kappa}}I_d$ and $h_{\epsilon}(O(\epsilon^2))=O(\epsilon^2)$.
In other words,
\begin{equation}
  \label{eq:smooth-filtered-covariance}
  h_{\epsilon}(C_{\epsilon}(x)) =
  \left(
    \begin{bmatrix}
      \frac{1}{p(x)m_{\kappa}} I_d & 0 \\
      0 & 0
    \end{bmatrix}
  \right)
  + O(\epsilon^2).
\end{equation}
Putting the two bounds together, this implies
\begin{equation}
  \label{eq:filtered-flux}
  h_{\epsilon}(C_{\epsilon}(x)) J(\nabla_{\epsilon}f(x))
  =
  m_{\kappa}^{-1}p^{-1}(x)\nabla f(x)
  + O(\epsilon).
\end{equation}

\subsection{Smoothed Finsler Laplacian}

We begin by bounding the difference between $\Delta[f](x)$ and $\Delta_{\epsilon}[f](x)$.
For fixed $x$, we apply a rigid transformation as before, so that $x=0$ and the manifold is a local graph of the tangent plane $y_x(u)=(u, g_x(u))$.
For convenience, define the vector field on the tangent plane
\begin{equation}
  \label{eq:normalized-flux}
  Q_x[f](u) = \frac{J(\nabla f(y_x(u)))}{p(x)}
\end{equation}
so that by \eqref{eq:filtered-flux} we have
\begin{equation}
  \label{eq:normalized-flux-2}
  h_{\epsilon}(C_{\epsilon}(y_x(u)))J(\nabla_{\epsilon} f(y_x(u)))
  =
  m_{\kappa}^{-1}
  \left(
    Q_x[f](u) + O(\epsilon)
  \right).
\end{equation}
Substituting \eqref{eq:normalized-flux-2} into \eqref{eq:smooth-fins-laplacian-2} and rewriting the integral as one over the tangent plane yields, after a change of variables,
\begin{equation}
  \label{eq:smooth-fins-laplacian-3}
  \Delta_{\epsilon}[f](0) =
  \frac{1}{m_{\kappa}\epsilon^{2}}
  \int
  \kappa\left(
    \frac{y(\epsilon v)}{\epsilon}
  \right)
  \left( Q_x[f](\epsilon v) + Q_x[f](0) + O(\epsilon) \right)\cdot
  y(\epsilon v)
  p_x(\epsilon v) dv.
\end{equation}
We perform a Taylor expansion on the product $(Q_x[f](\epsilon v)+Q_x[f](0) + O(\epsilon))p_x(\epsilon v)$, and applying the same parity arguments as before, simplify this expression to
\begin{equation}
  \label{eq:smooth-fins-laplacian-5}
  \begin{aligned}
    \Delta_{\epsilon}[f](x)
    &=
      \Tr\left(p_x(0)\nabla Q_x[f](0)
      + 2Q_x[f](0)\otimes\nabla p_x(0)\right)
      + O(\epsilon) \\
    &=
      2\frac{\langle J(\nabla f(x)), \nabla p(x)\rangle}{p(x)}
      + p(x)\div\left(\frac{J(\nabla f(x))}{p(x)}\right)
      + O(\epsilon) \\
    &=
      \frac{1}{p(x)}\div\left(p(x)J(\nabla f(x))\right) + O(\epsilon) \\
    &= \Delta[f](x) + O(\epsilon).
  \end{aligned}
\end{equation}
Thus, for fixed $x$ and $f$, $\Delta_{\epsilon}[f](x)$ is an $O(\epsilon)$-approximant of $\Delta[f](x)$.

\subsection{Concentration Inequalities}

We now bound the difference between $\Delta_{\epsilon}[f](x)$ and $\Delta_{\epsilon,n}[f](x)$ where $x$ is assumed to be a sampled point.
The expression for the empirical Laplacian \eqref{eq:disc-fins-laplacian-ptwise-2} closely resembles a Monte Carlo estimate of the smoothed Laplacian \eqref{eq:smooth-fins-laplacian-2}, with the exception of the smoothed gradients $\nabla_{\epsilon,n} f$ coupling the terms in the sum.
To account for this, we introduce an ``oracle'' empirical Laplacian that has access to the true (smoothed) gradients $\nabla_{\epsilon} f$, and then bound the difference in two steps via the triangle inequality.
In particular, defining
\begin{equation}
  \label{eq:disc-fins-laplacian-ptwise-oracle}
  \begin{aligned}
    \widetilde{\Delta}_{\epsilon,n}[f](x)
    = \frac{1}{n\epsilon^{d+2}}\sum_{i=1}^n
    \kappa\left(\frac{x-x_i}{\epsilon}\right)
    &\Big(
      h_{\epsilon}(C_{\epsilon}(x)) J(\nabla_{\epsilon} f(x)) \\
    &\quad+
      h_{\epsilon}(C_{\epsilon}(x_i)) J(\nabla_{\epsilon} f(x_i))
      \Big)
      (x-x_i)
  \end{aligned}
\end{equation}
we consider the bound
\begin{equation}
  \label{eq:disc-bound-strategy}
  \Big| \Delta_{\epsilon}[f](x) - \Delta_{\epsilon,n}[f](x) \Big|
  \leq
  \Big| \Delta_{\epsilon}[f](x) - \widetilde{\Delta}_{\epsilon,n}[f](x) \Big|
  + \Big| \widetilde{\Delta}_{\epsilon,n}[f](x) - \Delta_{\epsilon,n}[f](x) \Big|.
\end{equation}

\subsubsection{Sampling Error}

We consider the first term in the bound \eqref{eq:disc-bound-strategy}, which is the error due to discretization without the effects of gradient estimation.
Observe that $\widetilde{\Delta}_{\epsilon,n}[f](x_j)$ is an average of \textit{i.i.d.} terms bounded by $O(\epsilon^{-(d+1)})$, with variance $O(\epsilon^{-(d+2)})$, and whose expectation is equal to $\Delta_{\epsilon}[f](x)$.
Recalling that $J$ is assumed Lipschitz, Bernstein's inequality for bounded random variables yields the concentration inequality for fixed $x$:
\begin{equation}
  \label{eq:disc-oracle-bernstein}
  \mathbb{P}
  \Big\{
  |\Delta_{\epsilon}[f](x)-\widetilde{\Delta}_{\epsilon,n}[f](x)|
  <
  \frac{t}{\epsilon}
  \Big\}
  >
  1-2\exp\Big(
  -t^2n\epsilon^d/3
  \Big).
\end{equation}

\subsubsection{Covariance Estimation}

Since $C_{\epsilon,n}(x)$ is the average of $n$ \textit{i.i.d.} random matrices with operator norm uniformly bounded by $O(\epsilon^{-d})$ and expectation $C_{\epsilon}(x)$, the matrix Hoeffding inequality implies that for $t\geq 0$,
\begin{equation}
  \label{eq:disc-covariance-hoeffding}
  \mathbb{P}
  \Big\{
  \|C_{\epsilon,n}(x)-C_{\epsilon}(x)\| < t
  \Big\}
  >
  1-D\cdot\exp\left(-t^2n\epsilon^d/2\right).
\end{equation}
For $t=O(\epsilon^2)$, this implies
\begin{equation}
  \label{eq:filtered-disc-covariance-hoeffding}
  \mathbb{P}
  \Big\{
  \|h_{\epsilon}(C_{\epsilon,n}(x))-h_{\epsilon}(C_{\epsilon}(x))\| < t
  \Big\}
  >
  1-D\cdot\exp\left(-t^2n\epsilon^d/2\right).
\end{equation}

\subsubsection{Cogradient Estimation}

As a consequence of a previous approximation \eqref{eq:smooth-cograd-2}, we have
\begin{equation}
  \label{eq:smooth-cograd-norm-bound}
  \|b_{\epsilon}[f](x)\| \leq p(x)m_{\kappa}\|\nabla f(x)\| + O(\epsilon^2).
\end{equation}
For fixed $x$, the empirical cogradient $b_{\epsilon,n}[f](x)$ is the mean of $n$ \textit{i.i.d.} random vectors with norm uniformly bounded by $O(\epsilon^{-d})$, expected value $b_{\epsilon}[f](x)$, and variance bounded by $O(\epsilon^{-d})$.
The vector Bernstein inequality yields
\begin{equation}
  \label{eq:disc-cogradient-bernstein}
  \mathbb{P}
  \Big\{
  \|b_{\epsilon,n}[f](x)-b_{\epsilon}[f](x)\| < t
  \Big\}
  >
  1-2\exp\left(-t^2n\epsilon^d\right).
\end{equation}

\subsubsection{Gradient Approximation Error}

We now consider the second term in the bound \eqref{eq:disc-bound-strategy}, which is the error in the empirical Laplacian due to estimation of the gradients at the sampled points.
By a union bound over \eqref{eq:filtered-disc-covariance-hoeffding} and \eqref{eq:disc-cogradient-bernstein}, the empirical gradient satisfies
\begin{equation}
  \label{eq:disc-gradient-bound}
  \mathbb{P}
  \Big\{
  \|\nabla_{\epsilon,n}f(x)-\nabla_{\epsilon}f(x)\| < O(t)
  \Big\}
  >
  1-(2+D)\exp\left(-t^2n\epsilon^d/2\right),
\end{equation}
where we assume $t=O(\epsilon^2)$.
Since $J$ is assumed Lipschitz, these conditions imply our desired bound.
Namely,
\begin{equation}
  \label{eq:disc-flux-bound}
  \mathbb{P}
  \Big\{
  \|h_{\epsilon}(C_{\epsilon,n}(x))J(\nabla_{\epsilon,n}f(x))-
  h_{\epsilon}(C_{\epsilon}(x))J(\nabla_{\epsilon}f(x))\|
  < O(t)
  \Big\}
  >
  1-(2+D)\exp\left(-t^2n\epsilon^d/2\right).
\end{equation}

\subsection{Proof of \Cref{thm:graph-uniform-convergence}}

For convenience, define
\begin{equation}
  \label{eq:flux-shorthand}
  \begin{aligned}
    \Phi_{\epsilon}(x)
    &= h_{\epsilon}(C_{\epsilon}(x))J(\nabla_{\epsilon}f(x)) \\
    \Phi_{\epsilon,n}(x)
    &= h_{\epsilon}(C_{\epsilon,n}(x))J(\nabla_{\epsilon,n}f(x)).
  \end{aligned}
\end{equation}

For fixed index $i$, \eqref{eq:disc-flux-bound} coupled with the law of total probability implies
\begin{equation}
  \label{eq:flux-bound-xi}
  \mathbb{P}
  \Big\{
  |\Phi_{\epsilon}(x_i)-\Phi_{\epsilon,n}(x_i)|
  < O(t)
  \Big\}
  >
  1-(2+D)\exp\left(-t^2(n-1)\epsilon^d/2\right).
\end{equation}
By a union bound over the indices $i$, we have
\begin{equation}
  \label{eq:flux-bound-all-xi}
  \mathbb{P}
  \Big\{
  \max_i|\Phi_{\epsilon}(x_i)-\Phi_{\epsilon,n}(x_i)|
  < O(t)
  \Big\}
  >
  1-(2+D)n\exp\left(-t^2(n-1)\epsilon^d/2\right).
\end{equation}
These bounds taken together imply
\begin{equation}
  \label{eq:disc-laplacian-err-1}
    |\widetilde{\Delta}_{\epsilon,n}[f](x_j) - \Delta_{\epsilon,n}[f](x_j)|
    = O(t\epsilon^{-(d+1)})
\end{equation}
for all indices $1\leq j\leq n$ with probability at least $1-(2+D)n\exp(-t^2(n-1)\epsilon^d/2)$.
Taken together with \eqref{eq:disc-oracle-bernstein}, we have
\begin{equation}
  \label{eq:disc-laplacian-err-2}
  |\Delta_{\epsilon}[f](x_j) - \Delta_{\epsilon,n}[f](x_j)|
  = O(t\epsilon^{-(d+1)})
\end{equation}
with probability at least $1-(n(2+D)+1)\exp(-t^2(n-1)\epsilon^d/2)$.
We choose $t,\epsilon$ varying with $n$ according to
\begin{equation}
  \label{eq:constant-choices}
  \begin{gathered}
    \epsilon_n = \left( (4+2\alpha) \frac{\log n}{n} \right)^{\frac{1}{3d+4}} \\
    t_n = \epsilon_n^{d+2}
  \end{gathered}
\end{equation}
for some $\alpha>0$, so that
\begin{equation}
  \label{eq:disc-laplacian-err-2}
  \mathbb{P}
  \left\{
    \max_j|\Delta_{\epsilon}[f](x_j) - \Delta_{\epsilon,n}[f](x_j)|
    < O(\epsilon_n)
  \right\}
  >
  1-O \left( \frac{D}{n^{1+\alpha}} \right).
\end{equation}
Then, by the bound \eqref{eq:smooth-fins-laplacian-5}, we have
\begin{equation}
  \label{eq:disc-laplacian-err-3}
  \mathbb{P}
  \left\{
    \max_j|\Delta[f](x_j) - \Delta_{\epsilon,n}[f](x_j)|
    < O(\epsilon_n)
  \right\}
  >
  1-O \left( \frac{D}{n^{1+\alpha}} \right).
\end{equation}
Noting that the empirical estimates are assumed independent, applying the Borel-Cantelli Lemma establishes \Cref{thm:graph-uniform-convergence}, as desired.

\section{Implementation Details}
\label{sec:methods}

In this section, we detail the specifics of the experiments in \Cref{sec:experiments}.
All experiments are implemented in Python using the JAX, Equinox, Optax, and Diffrax libraries~\citep{jax2018,deepmind2020,kidger2021a,kidger2021b}.

\subsection{Convergence on the Finsler Graph Laplacian}

For a torus with major radius $R=2.0$ and minor radius $r=1.0$, points are sampled via rejection sampling.
Azimuthal and poloidal angles $(u, v)$ are uniformly proposed, and accepted according to the true area (Hausdorff) measure $dA = r(R + r\cos(v)) du dv$.
We compute $C_{\epsilon,n}^\dagger(x)$ by taking the Moore-Penrose pseudoinverse of the best rank-$2$ approximant of the local covariance matrix $C_{\epsilon, n}(x)$.

\subsection{Learning Geometry using Finslerian Graph Neural Networks}

The initial conditions $f(0)$ are generated as a superposition of 5 random Fourier modes with frequencies sampled uniformly from $[1.0, 3.0]$ and phases from $[0, 2\pi]$.
The drift vector defining the Randers metric is set to $v = [0.25, 0.25]$.
The parameterizations of $\hat{J}(\xi)$ are constructed using a neural network with a single hidden layer of with $E=64$.
The graph neural diffusion equations are integrated from $t=0$ to $t=1$ using the Tsitouras 5/4 solver~\citep{tsitouras2011}, with a time increment of $\Delta t=0.05$.
Models are trained using the Adam optimizer with a learning rate of $3\times 10^{-4}$ for $1000$ steps, with weight decay.
The loss function is the relative Mean Squared Error (MSE) computed exclusively over 20\% of the nodes selected uniformly at random.

\end{document}